\documentclass[conference]{IEEEtran}
\usepackage{cite}
\usepackage{amsmath,amssymb,amsfonts}
\usepackage{algorithmic}
\usepackage{graphicx}
\usepackage{textcomp}
\usepackage[english]{babel}
\usepackage{amsmath}
\usepackage{graphicx,epstopdf}
\usepackage[colorlinks=true, allcolors=blue]{hyperref}
\usepackage[english]{babel}
\usepackage[ruled,vlined]{algorithm2e}
\usepackage{amsmath}
\usepackage{booktabs}
\usepackage{hyperref}
\usepackage{graphicx}
\usepackage{amsmath,amssymb}
\usepackage{latexsym}
\usepackage{eurosym}
\usepackage{mathtools}
\usepackage{lscape} 
\usepackage{amssymb}
\usepackage{xcolor}
\usepackage{lipsum}
\usepackage{optidef}
\usepackage{multirow}
\usepackage{booktabs}
\usepackage{siunitx}

\def\BibTeX{{\rm B\kern-.05em{\sc i\kern-.025em b}\kern-.08em
    T\kern-.1667em\lower.7ex\hbox{E}\kern-.125emX}}

\usepackage{svg}
\IEEEoverridecommandlockouts
\title{Optimizing Vehicular Networks with Variational Quantum Circuits-based Reinforcement Learning \vspace{-5mm} }
\author{
  \IEEEauthorblockN{%
    Zijiang~Yan\IEEEauthorrefmark{1}, 
    Ramsundar~Tanikella\IEEEauthorrefmark{2},
    Hina~Tabassum\IEEEauthorrefmark{1} \thanks{This work  was
supported by a Discovery Grant funded by the Natural Sciences and
Engineering Research Council of Canada and MITACS. At the time of this work, R. Tanikella was with the Department of Electrical Engineering and Computer Science at York University, Toronto,  Canada.}
  }%
  \IEEEauthorblockA{\IEEEauthorrefmark{1}York University, Canada,%  
  \IEEEauthorrefmark{2}Indian Institute of Technology, Bhubaneswar, India \vspace{-5mm} }
  }
  
\begin{document}

\maketitle
\begin{abstract}
In vehicular networks (VNets), ensuring both road safety and dependable network connectivity is of utmost importance. Achieving this necessitates the creation of resilient and efficient decision-making policies that prioritize multiple objectives. In this paper, we develop a Variational Quantum Circuit (VQC)-based multi-objective reinforcement learning (MORL) framework to characterize efficient network selection and autonomous driving policies in a  vehicular network (VNet).  Numerical results showcase notable enhancements in both convergence rates and rewards when compared to conventional deep-Q networks (DQNs), validating the efficacy of the VQC-MORL solution. 
\raggedbottom
% Numerical results show a comparison between the classical RL frameworks like Q-learning, Deep Q-Networks (DQN) and QRL frameworks. 

% In this paper, we develop a Quantum inspired reinforcement learning (QRL) framework to characterize efficient network selection and autonomous driving policies in a multi-band vehicular network (VNet) operating on conventional Radio frequencies (RF) and Terahertz (THz) frequencies. The framework we introduce is strategically devised to achieve two primary objectives. Firstly, it focuses on optimizing traffic flow while minimizing collisions by exerting control over the vehicle's motion dynamics, encompassing aspects such as speed and acceleration, from the standpoint of autonomous driving. Secondly, it strives to enhance data rates and reduce HOs by synchronously managing the vehicle's motion dynamics and network selection, bridging the realms of telecommunications and vehicular dynamics.

\end{abstract}

\section{Introduction}
Jointly optimizing  the kinematics and network connectivity of autonomous  vehicles (AVs) is imperative to mitigate road collisions. However, the highly stochastic nature of the wireless transmission channels and road traffic necessitate faster decision making. Compared to traditional optimization, reinforcement learning (RL) offers robust and fast decision-making for AVs. However, in practice, often RL is susceptible to  high dimensional state-action spaces and a huge training data; thus,  classical RL becomes time consuming and not scalable.   Recently, variational quantum circuits (VQCs) have been shown to offer better trade-off between exploration and exploitation compared to classical RL methods.

% contribution
We investigate the performance of a VQC-based RL method to optimize both cell-association and autonomous driving policies on a multi-lane highway equipped with base-stations (BSs) operating on RF and THz spectrum. The objective is to maximize handoff (HO)-aware data rates and traffic flow while ensuring collision avoidance. We formulate the problem as a multi-objective Markov decision process (MOMDP) and subsequently convert this  into quantum eigen-states and eigen-actions using quantum circuits. The proposed VQC method strategically employs quantum circuits in lieu of conventional neural networks.  The proposed VQC-MORL method outperforms traditional deep-Q network (DQN) in terms of convergence and obtained rewards.

\section{System Model and Assumptions}
We consider a two-tier downlink network composed of $N_R$ RF BSs (RBSs) and $N_T$ THz BSs (TBSs) in a multi-vehicle environment on a four-lane road. Here, there are $V$ AVs that receive data from these roadside BSs \cite{yan2022gc}. Each AV selectively connects to a single BS (either RBS or TBS). The AVs' on-board units (OBUs) collect real-time data on VNet, encompassing neighboring vehicles' velocity, acceleration, and lane positions. We model individual driver behavior  as in \cite{kesting2010enhanced}. The signal-to-interference-plus- noise ratio (SINR) for $j$-th AV  from BS $i$ is modeled as in \cite{yan2022gc}. In THz network, the SINR of a $j$-th AV is modeled as \cite{mobility}.  All AVs are equipped with a single antenna. We assume the typical AV's receiving beam aligns with the transmitting beam of the associated TBS through beam alignment techniques. However, the alignment of the main lobe of  the user and interfering TBSs is modeled with the probability $q$. The interference with this probability is computed as in \cite{mobility}.  Each RBS and TBS has the available bandwidth given by $W_R$ and $W_T$, respectively.  The data rate of each AV to BS link  can be computed as $ R_{ij} = W_j\log_2(1+\mathrm{SINR}_{ij})$.
Each RBS and TBS has a maximum limit of $Q_R$ and  $Q_T$ on the number of AVs that can be supported, respectively.  Each AV maintains a set of top three BSs in terms of the achievable data rate  if their respective $\mathrm{SINR}_{ij}(t)\geq \gamma_{th}$.  Consequently, each BS can calculate the number of possible AVs at each time instance denoted by $n_i$. As AVs drive along the corridor, they switch from (connecting to) one BS to another when $\mathrm{SINR}_{ij}< \gamma_{th}$.  Frequent HOs can reduce the data rate due to HO latency failures.   We discourage HOs by introducing a HO penalty ($\mu$) which is higher for TBS and lower for RBS.

\section{MOMDP Formulation and VQC-MORL}

\subsubsection{MOMDP Formulation}

Our observation space is composed of driving and communication observations.  The driving observation space  is included in the \textit{highway-env}  environment \cite{highway-env}. 
Consequently, a state $s_t$ for AV $j$ constitutes location of AVs, velocity of AVs, number of AVs that associates with a BS $i$, SINR of AVs with BSs,  i.e.,
$s_t=( \textbf{q}_j(t), \textbf{v}_j(t),  n_i(t),  \text{SINR}_{ij}(t))$. At each time step $t$, AV $j$ selects action $a_t=(a_t^{\rm tran}, a_t^{ \rm tele}) \in \mathcal{A}_{\rm tran} \times \mathcal{A}_{\rm tele}$, where $a_t^{\rm tran}$ is the driving action, i.e., trajectory action and $a_t^{tele}$ is the communication-related action, i.e., association with a BS.
$\mathcal{A}_{\rm tran}=\{a_{\rm tran}^1,\ldots,a_{\rm tran}^5\}$, where $a_{\rm tran}^1$ is the change lane to the left lane action, $a_{\rm tran}^2$ is maintaining the same lane, $a_{\rm tran}^3$ is the change lane to the right lane action, $a_{\rm tran}^4$ is accelerating within the same lane, and $a_{\rm tran}^5$ is decelerating within the same lane. Similarly, the communication action space at time $t$ can be given by $\mathcal{A}_{\rm tele}=\{ a_{\rm tele}^1,a_{\rm tele}^2,a_{\rm tele}^3 \}$. In $a_{\rm tele}^1,$, AV selects a BS with maximum \textit{weighted rate metric} that encourages traffic load balancing between BSs and discourages unnecessary HOs, especially for TBSs, i.e.,
$
    \text{WR}_{ij}(t) = \frac{R_{ij}(t)}{\min \left(Q_i, n_i(t) \right)} (1 - \mu), \forall i.
$
In $a_{\rm tele}^2,$ the AV selects a BS with maximum $\text{WR}_{{ij}}$ by substituting $\mu=0$, if $Q_i \geq n_i(t)$. Otherwise, AV recursively selects the next vacant best-performing BS in terms of $\text{WR}_{ij}$. 
\textbf{(3)} In $a_{\rm tele}^3$, the AV  chooses to connect to a BS with the maximum  data rate $\text{R}_{ij}$.
We define the AV driving reward as follows \cite{highway-env} and \cite{yan2023multi}:
\begin{equation}
    r^{\mathrm{tran}}_{j} (t)=  \omega_1 \left( \frac{||\mathbf{v}_j(t)|| -v_\mathrm{min}}{v_\mathrm{max}-v_\mathrm{min}} \right)- \omega_2 \cdot \delta, \forall k \in \mathcal{U},
\end{equation}
where $\mathbf{v}_j(t),v_{\min}$ and $v_{\max}$ are the current longitudinal velocity for AV $j$ on timestep $t$, the minimum and maximum speed limits, and $\delta$ is the collision indicator. $\omega_1$ and $\omega_2$ are the weights that adjust the value of the AV driving reward with its collision penalty. We define the communication reward as $
    r^{\mathrm{tele}}_{j}(t)= \omega_3 R_{i_0k}(t) \left(1- \text{min}(1,\xi_k (t))\right),% + w_6 j +w_5 l
$
where  $R_{i_0k}(t)$ is the achievable data rate when associated with BS $i_0$, and  $\xi_k(t)$ is the HO probability, computed by dividing the number of HOs accounted until the current time $t$ by the time duration of previous time slots in the episode.

\subsubsection{Proposed VQC-MORL Solution}

We use VQC as a Q-function approximator instead of a neural network. 
\begin{algorithm}
\footnotesize
\SetAlgoLined
\KwResult{Quantum Circuit $U_\theta$  }
\KwData{ Quantum Circuit $U_\theta$, Experience replay memory $D$, mini batch-size $m$}
\textbf{Initialization:} $\mathcal{D} \gets \bold{0}$, $\theta \gets \bold{0}$, Target quantum circuits $\theta^* \gets \theta$, RBSs, TBSs, AVs \\
\While{$\mathrm{episode} < \mathrm{episode \ limit }$}{
  $t \gets 0$, $s_1$ initial and encode it to quantum state\\
\While{$t \leq \mathrm{horizon \ limit} $}{AV selects $a_t$ by $\epsilon$-greedy search as  
% \textbf{Algorithm-1}.\\  
% \begin{align*}
% a_t
% \begin{dcases} \text{Select from } \mathcal{A} & \text{probability of } $\epsilon$ \\
% \text{Select }a_t \max_{a \in \mathcal{A}}{Q_{\theta}(s_t,a_t,\theta)} & \text{probability of } $1 - \epsilon$
% \end{dcases}
% \end{align*}
% $a_t$ select action from $\mathcal{A}$ with probability of $\epsilon$ or\\  Select $a_t$ from $\max_{a \in \mathcal{A}}{Q_{\theta}(s_t,a_t,\theta)}$ with probability of $ 1-\epsilon$; \\
 $ a^{\mathrm{tele} }_{t}$ and $a^{\mathrm{tran} }_{t}$  
% Derive $ a^{\mathrm{tele} }_{t}$ and $a^{\mathrm{tran} }_{t}$  from $a_t$;\\
and Enforce $a^{\mathrm{tele} }_{t}$ and $a^{\mathrm{tran} }_{t}$ to AV;\\
% Calculate reward $r_t$ and update $s'$;\\
% Store $(s_t,a_t,s_{t+1},r_t)$ to $\mathcal{D}$;\\
\textbf{Experience Replay:} sample mini-batch transitions in $\mathcal{D}$ $(s_k,a_k,r_k,s'_k)$ where  $k \in m$;\\
\textbf{Set target-$Q$ function:}  \\ 
% $Q(s,a)= <0^{n}| U_{\theta}^+(s0O_a$
$Q(s,a;\theta) = \langle O_a \rangle _{s,{\theta}}$ \\

\textbf{Set real $Q$-function:} $  {Q}(s_t,a_t;\theta)$ by Eq.\ref{eq:vqc-bellman}\\

Compute loss: $\mathcal{L}(\theta) $ by Eq.\ref{eq:vqc-loss} \\

Perform gradient descent step by minimizing loss $\mathcal{L}(\theta)$;
$\theta \gets \theta - a_{t} \cdot \mathcal{L}(\theta) \cdot \triangledown_{\theta}{y_k}$;\\

Update the $U_\theta$ weights $\theta \gets \theta^*$;
% $Q_d(s,a_d) = V(s) + (A_d(s,a_d) - \max_{a'_d \in A_d}A_d(s,a'_d))$
% $Q_d(s,a_d) = (A_d(s,a_d) - \max_{a'_d \in A_d}A_d(s,a'_d))$
  }
   Policy updated in terms of $U_\theta$
 }
  \caption{VQC-MORL Algorithm}
\end{algorithm}
% VQCs take as input the agent's state (i.e., a numpy array of features) and outputs a vector of expectation values. These expectation values are then post-processed to approximate the Q-values. VQC encodes the input vector through the use of single-qubit rotations, where rotation angles are controlled by the components of this input vector. In order to get a highly-expressive model, these single-qubit encodings are not performed only once in the VQC, but in several "re-uploadings", interlayed with variational gates \cite{fakhari2013quantum}. 
As opposed to a policy-gradient approach , VQCs approximate the $Q$-function of the agent by defining a function approximator given by
\begin{equation}
    Q(s,a;\theta) = \langle O_a \rangle _{s,{\theta}} = \langle 0^{\otimes 5} | U_\theta^\dagger(s) O_a U_\theta(s) | 0^{\otimes 5} \rangle
    \label{eq:vqc-bellman}
\end{equation}
where  $\langle O_a \rangle _{s,{\theta}}$  is expectation values (or the Q-values) of observables (the input features), $O_a$  (one per action) measured at the output of the VQC and weight by Pauli products, calculated using the value function given by  Eq. \ref{eq:vqc-bellman}. $U_\theta(s)$ is VQC which parametrized by $\theta$.  All observables are adjusted so that their expected values fall within the real number range, whose expectation fall in real number set as $\mathbb{E}(O_a) \in \mathbb{R}$.

%$V(s) = \sum\limits_{a \in \mathcal{A}_{s}} \pi (s,a)[R(s,a) + \gamma \sum\limits_{s'}P(s')V(s')]$

 %$V(s) \longleftarrow V(s)  + \alpha 
% (\gamma V(s') - V(s) + r)$
% \begin{equation}
% \label{eq:vqc-bellman}
%     V(s) \longleftarrow V(s)  + \alpha (\gamma V(s') - V(s) + r)
% \end{equation}
These updated $Q$-values are then passed into the loss function derived from $Q$-learning:
\begin{equation}
\label{eq:vqc-loss}
    \mathcal{L}(\theta) = \frac{1}{|D|} \sum\limits_{(s,a,r,s')\in D} (Q(s,a;\theta) - [r + \max_{a'} Q(s',a';\theta')])^{2}
\end{equation}

for a batch  $D$  of  1-step interactions  $(s,a,r,s')$  with the environment, sampled from the replay memory, and parameters  $\theta^{'}$  specifying the target VQC parameters. Gradient descent step on the above loss function gives the optimum $\theta$ values thus giving the most optimal action combination for a given state. 
Then, we defined DQN for this environment, we use $\epsilon-$ greedy defined policy agent. defined by $\pi(a|s) =  r_k + \gamma\max_{a'}\hat{Q}(s'_k,a'_k;\theta_k)$ with probability of $1 - \epsilon$, otherwise is $\pi(a|s) = 1/15$  where $\epsilon$ is decayed in each episode.
% \begin{equation}
%     \pi(a|s) = 
% \begin{dcases}
%      r_k + \gamma\max_{a'}\hat{Q}(s'_k,a'_k;\theta_k) ,& 1 - \epsilon \\
%     \frac{1}{15} , & \text{o.w.} \\
%     % 0, & \text{Keep Previous BS} 
% \end{dcases}
% \end{equation}

The operation of the proposed VQC-based RL approaches are summarized within \textbf{Algorithm 1}.

\section{Numerical Results and Discussions}
\begin{figure}[h]
\centering
\begin{tabular}{lccccc}
\includegraphics[width=0.3\linewidth]{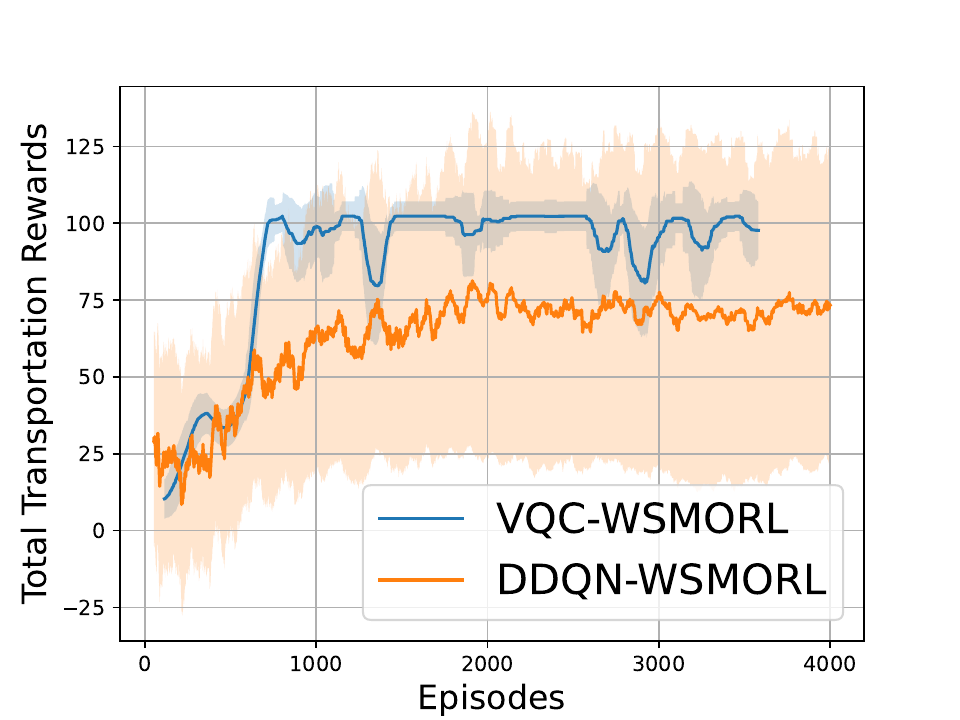}\hspace{-1cm}&
\includegraphics[width=0.3\linewidth]{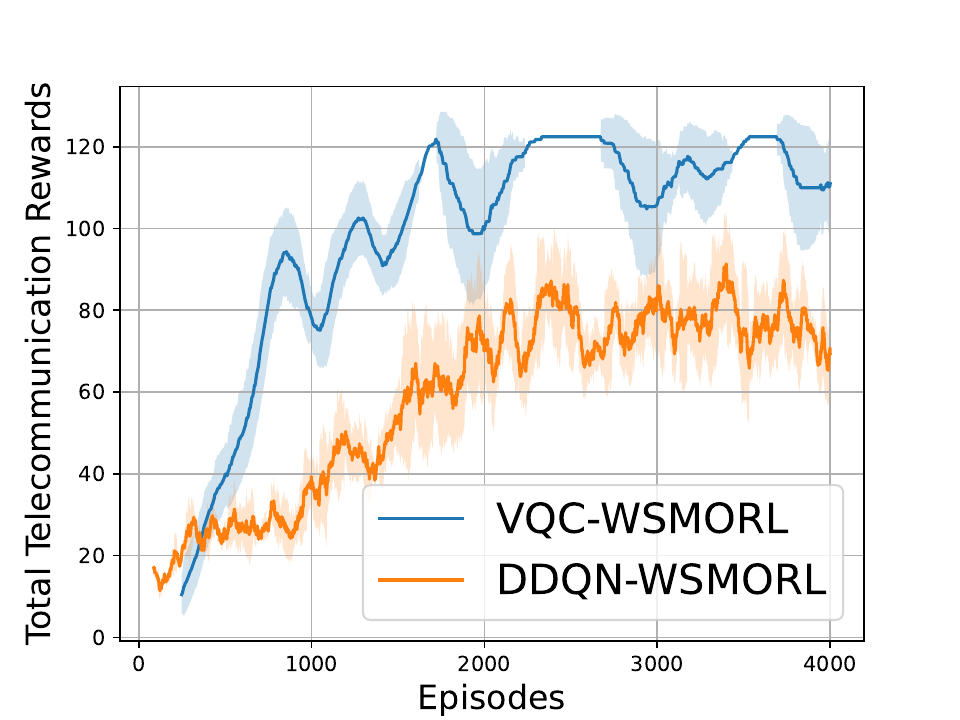}\hspace{-1cm}&
\includegraphics[width=0.3\linewidth]{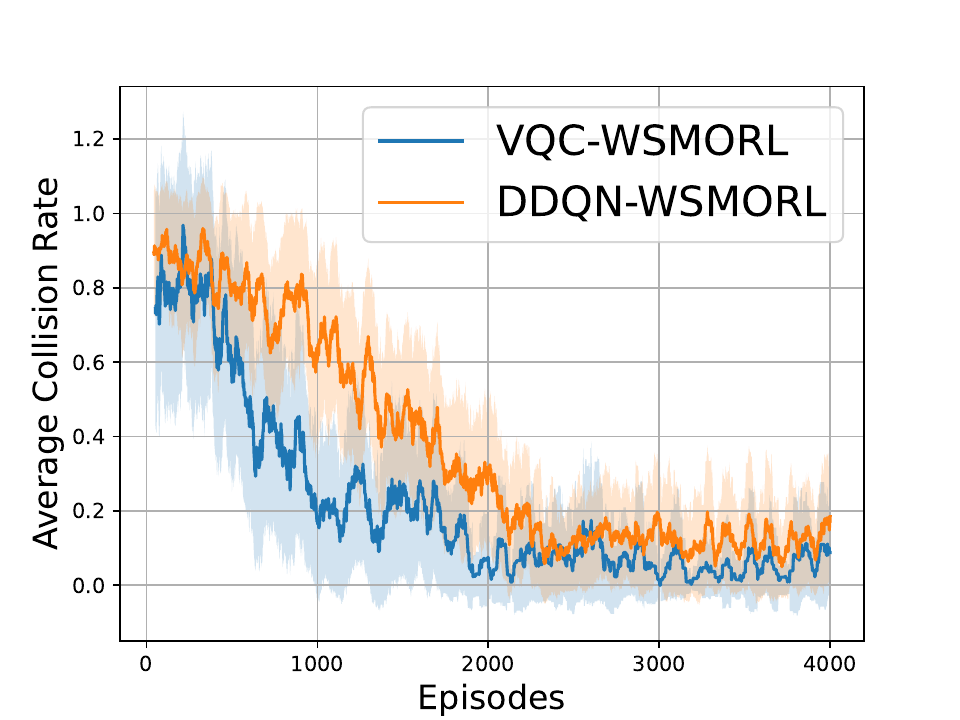}\hspace{-1cm}&
&\\
\qquad (a)  & (b) & (c)
\end{tabular}
\caption {Training performances (ego vehicle): (a) Total telecommunication reward (b) Total transport reward (c) Collision Rate}
\label{fig:training}
\end{figure}

Fig. \ref{fig:training} depicts that the training convergence speed of the proposed VQC-MORL algorithm outperforms conventional DDQN.  In contrast to the other benchmarks, VQC improves the training efficiency by $31.32\%$. Fig.2 depicts that leveraging VQC for Q-value representation proves advantages as it offers higher telecommunication and transportation rewards compared to  DDQN (average $18.64\%$ gain). 

\begin{figure}[h]
\centering
\begin{tabular}{lccccc}
\includegraphics[width=0.3\linewidth]{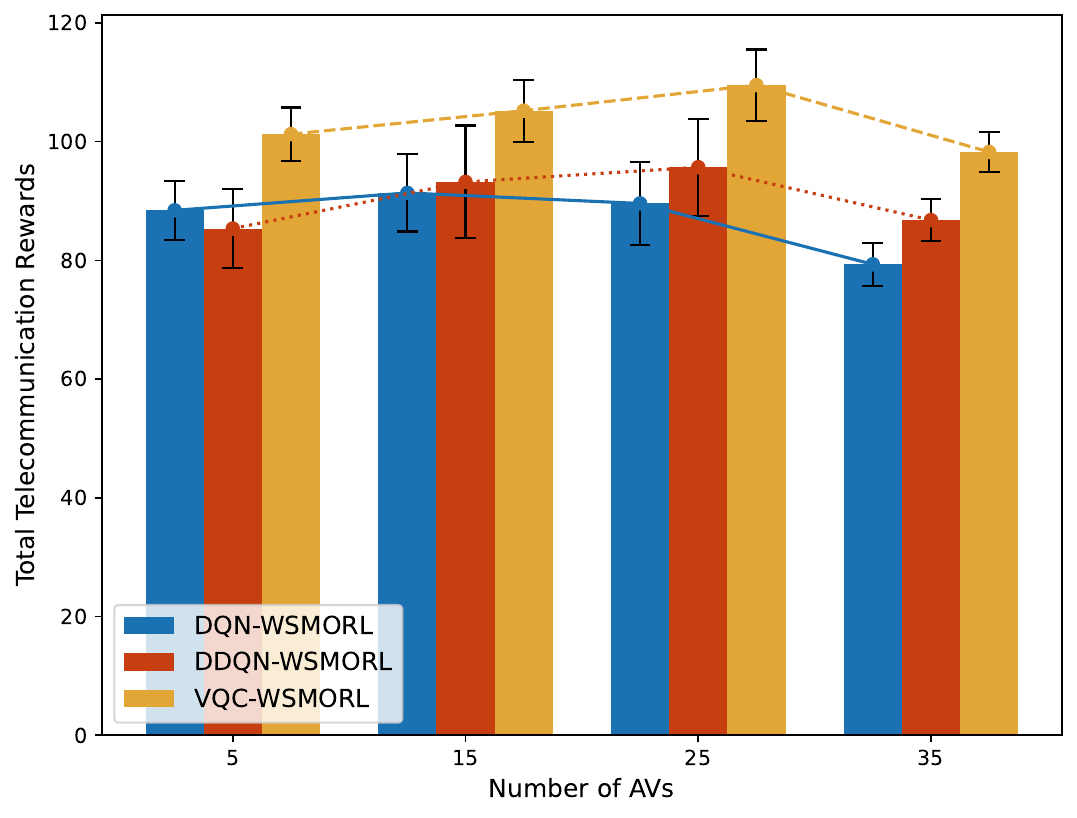}\hspace{-1cm}&
\includegraphics[width=0.3\linewidth]{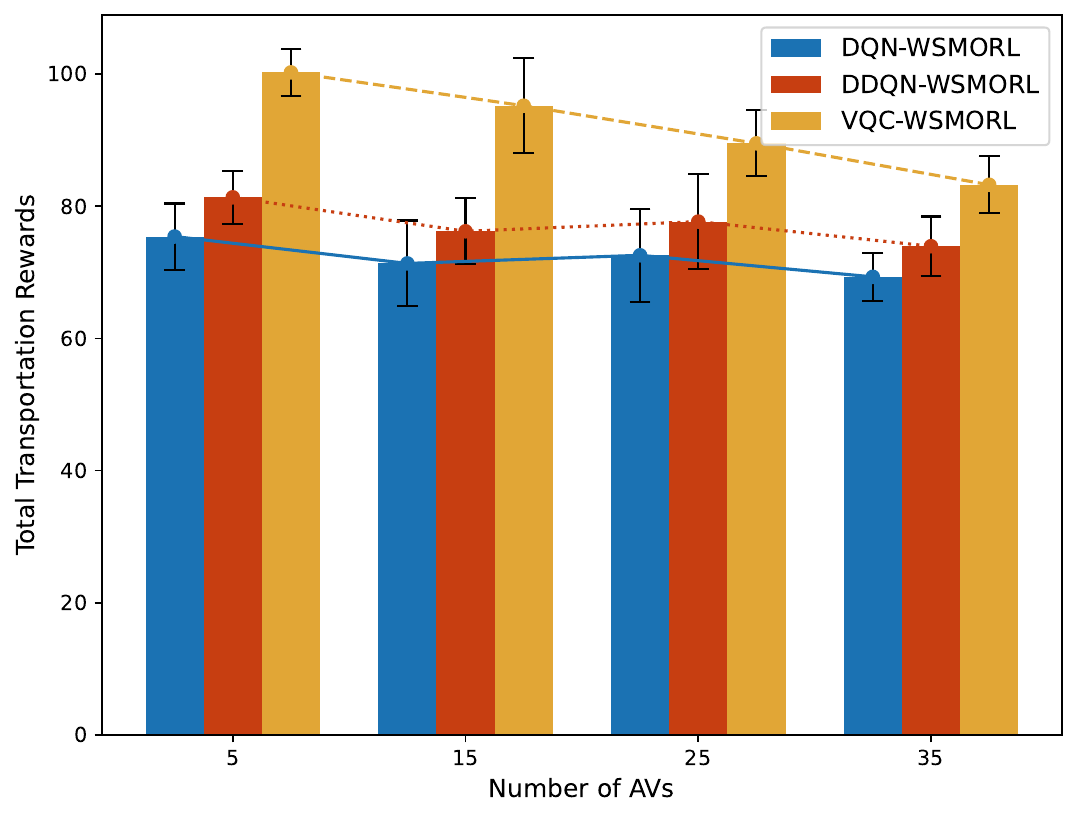}\hspace{-1cm}&
\includegraphics[width=0.3\linewidth]{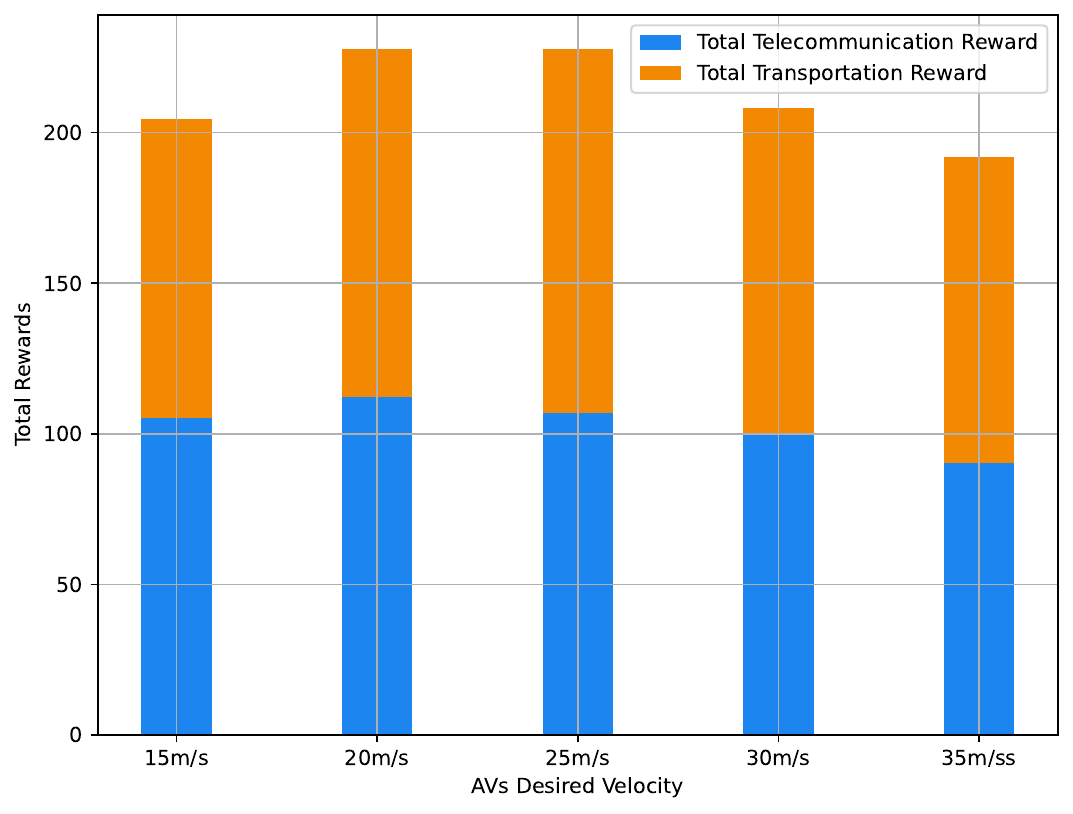}\hspace{-1cm}&
&\\
\qquad (a)  & (b) & (c)
\end{tabular}
\vspace{-7mm}
\caption {Testing performance (ego vehicle): (a) Total telecommunication reward (b) Total transport reward (c) Total reward. The considered VQC  architecture has 5 qubits and 3 layers.} 
\label{fig:evaluation}
\end{figure}

According to Fig. \ref{fig:evaluation}(a) and (b), introducing more AVs contributes high wireless resource sharing competition and high traffic flow. From Fig. \ref{fig:evaluation}(c), increasing the desired velocities  increases   total rewards initially since transportation rewards benefits exceed the handover loss on telecommunication rewards. More collisions and hanodvers occur with increasing velocities, reduce the total rewards.

\vspace{-3mm}
\vspace{-7mm}

\tiny
\bibliographystyle{IEEEtran}
\bibliography{ref.bib}

\begin{thebibliography}{1}
\providecommand{\url}[1]{#1}
\csname url@samestyle\endcsname
\providecommand{\newblock}{\relax}
\providecommand{\bibinfo}[2]{#2}
\providecommand{\BIBentrySTDinterwordspacing}{\spaceskip=0pt\relax}
\providecommand{\BIBentryALTinterwordstretchfactor}{4}
\providecommand{\BIBentryALTinterwordspacing}{\spaceskip=\fontdimen2\font plus
\BIBentryALTinterwordstretchfactor\fontdimen3\font minus
  \fontdimen4\font\relax}
\providecommand{\BIBforeignlanguage}[2]{{%
\expandafter\ifx\csname l@#1\endcsname\relax
\typeout{** WARNING: IEEEtran.bst: No hyphenation pattern has been}%
\typeout{** loaded for the language `#1'. Using the pattern for}%
\typeout{** the default language instead.}%
\else
\language=\csname l@#1\endcsname
\fi
#2}}
\providecommand{\BIBdecl}{\relax}
\BIBdecl

\bibitem{yan2022gc}
Z.~Yan and H.~Tabassum, ``Reinforcement learning for joint v2i network
  selection and autonomous driving policies,'' in \emph{GLOBECOM 2022 - 2022
  IEEE Global Communications Conference}, 2022, pp. 1241--1246.

\bibitem{kesting2010enhanced}
A.~Kesting, M.~Treiber, and D.~Helbing, ``Enhanced intelligent driver model to
  access the impact of driving strategies on traffic capacity,''
  \emph{Philosophical Transactions of the Royal Society A: Mathematical,
  Physical and Engineering Sciences}, vol. 368, no. 1928, pp. 4585--4605, 2010.

\bibitem{mobility}
M.~T. Hossan and H.~Tabassum, ``Mobility-aware performance in hybrid rf and
  terahertz wireless networks,'' \emph{IEEE Transactions on Communications},
  vol.~70, no.~2, pp. 1376--1390, 2022.

\bibitem{highway-env}
E.~Leurent, ``An environment for autonomous driving decision-making,''
  \url{https://github.com/eleurent/highway-env}, 2018.

\bibitem{yan2023multi}
Z.~Yan, W.~Jaafar, B.~Selim, and H.~Tabassum, ``Multi-uav speed control with
  collision avoidance and handover-aware cell association: Drl with action
  branching,'' in \emph{GLOBECOM 2023 - 2023 IEEE Global Communications
  Conference}, 2023, pp. 5067--5072.

\end{thebibliography}
\end{document}